\documentclass[sigconf,natbib=true,anonymous=false]{acmart}

\AtBeginDocument{%
  }

\setcopyright{acmlicensed}
\copyrightyear{2024}
\acmYear{2024}
\setcopyright{acmlicensed}\acmConference[SIGIR '24]{Proceedings of the 47th International ACM SIGIR Conference on Research and Development in Information Retrieval}{July 14--18, 2024}{Washington, DC, USA}
\acmBooktitle{Proceedings of the 47th International ACM SIGIR Conference on Research and Development in Information Retrieval (SIGIR '24), July 14--18, 2024, Washington, DC, USA}
\acmDOI{10.1145/3626772.3657706}
\acmISBN{979-8-4007-0431-4/24/07}

\usepackage{multirow} 
\usepackage{subcaption}

\begin{document}

\title{Transformer-based Reasoning for Learning Evolutionary Chain of Events on Temporal Knowledge Graph}

\author{Zhiyu Fang}
\authornote{Both authors contributed equally to this research.}
\email{mr.fangzy@foxmail.com}
\author{Shuai-Long Lei}
\authornotemark[1]
\email{shuailong0lei@gmail.com}
\affiliation{%
  \institution{University of Science and Technology Beijing}
  \city{Haidian}
  \state{Beijing}
  \country{China}
}

\author{Xiaobin Zhu}
\authornote{Corresponding author}
\affiliation{%
  \institution{University of Science and Technology Beijing}
  \city{Haidian}
  \state{Beijing}
  \country{China}
}
\email{zhuxiaobin@ustb.edu.cn}

\author{Chun Yang}
\affiliation{%
  \institution{University of Science and Technology Beijing}
  \city{Haidian}
  \state{Beijing}
  \country{China}
}
\email{chunyang@ustb.edu.cn}

\author{Shi-Xue Zhang}
\affiliation{%
  \institution{University of Science and Technology Beijing}
  \city{Haidian}
  \state{Beijing}
  \country{China}
}
\email{zhangshixue111@163.com}

\author{Xu-Cheng Yin}
\affiliation{%
  \institution{University of Science and Technology Beijing}
  \city{Haidian}
  \state{Beijing}
  \country{China}
}
\email{xuchengyin@ustb.edu.cn}

\author{Jingyan Qin}
\affiliation{%
  \institution{University of Science and Technology Beijing}
  \city{Haidian}
  \state{Beijing}
  \country{China}
}
\email{qinjingyanking@foxmail.com}

\renewcommand{\shortauthors}{Fang et al.}

\begin{abstract}
Temporal Knowledge Graph (TKG) reasoning often involves completing missing factual elements along the timeline. Although existing methods can learn good embeddings for each factual element in quadruples by integrating temporal information, they often fail to infer the evolution of temporal facts. This is mainly because of (1) insufficiently exploring the internal structure and semantic relationships within individual quadruples and (2) inadequately learning a unified representation of the contextual and temporal correlations among different quadruples. To overcome these limitations, we propose a novel Transformer-based reasoning model (dubbed ECEformer) for TKG to learn the Evolutionary Chain of Events (ECE). Specifically, we unfold the neighborhood subgraph of an entity node in chronological order, forming an evolutionary chain of events as the input for our model. Subsequently, we utilize a Transformer encoder to learn the embeddings of intra-quadruples for ECE. We then craft a mixed-context reasoning module based on the multi-layer perceptron (MLP) to learn the unified representations of inter-quadruples for ECE while accomplishing temporal knowledge reasoning. In addition, to enhance the timeliness of the events, we devise an additional time prediction task to complete effective temporal information within the learned unified representation. Extensive experiments on six benchmark datasets verify the state-of-the-art performance and the effectiveness of our method.
\end{abstract}




\maketitle

\section{Introduction}

\begin{figure}[t]
\centering
\begin{subfigure}{\linewidth}
    \centering
    \includegraphics[width=0.9\linewidth]{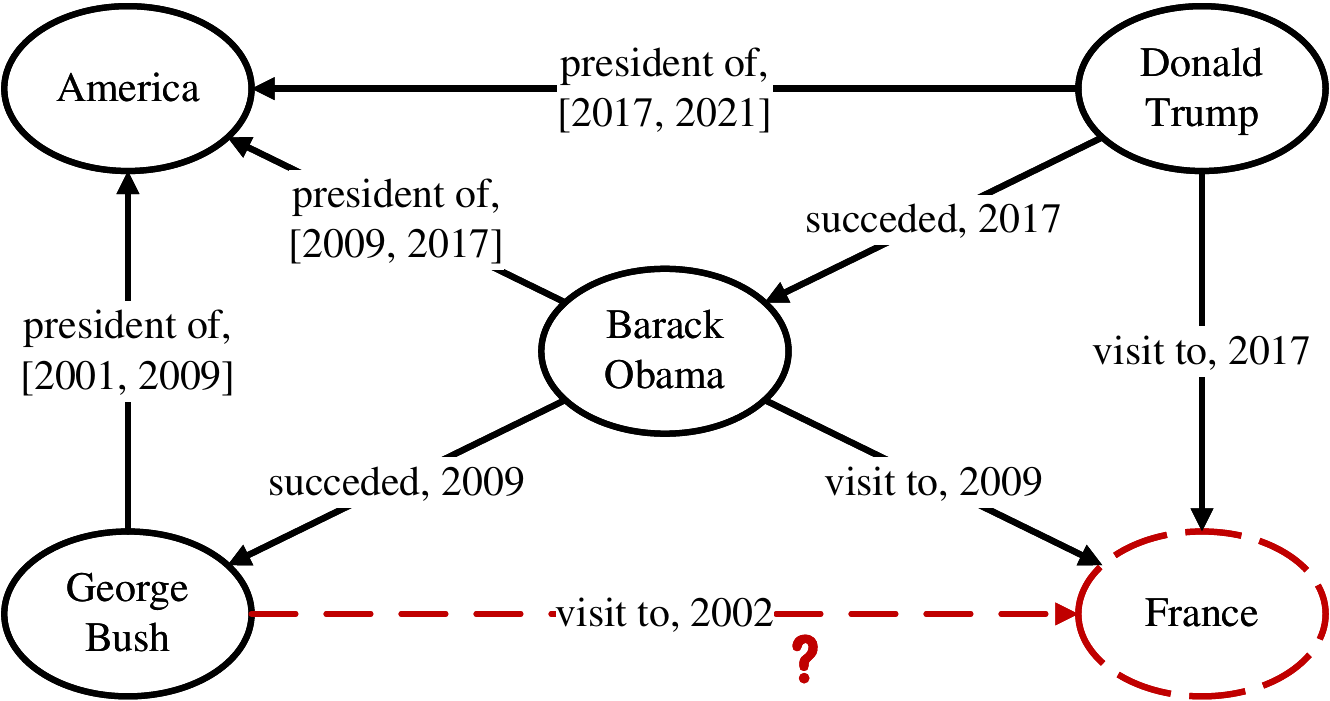} 
    \caption{}
    \label{fig:overall1}
\end{subfigure}
\begin{subfigure}{\linewidth}
    \centering
    \includegraphics[width=\linewidth]{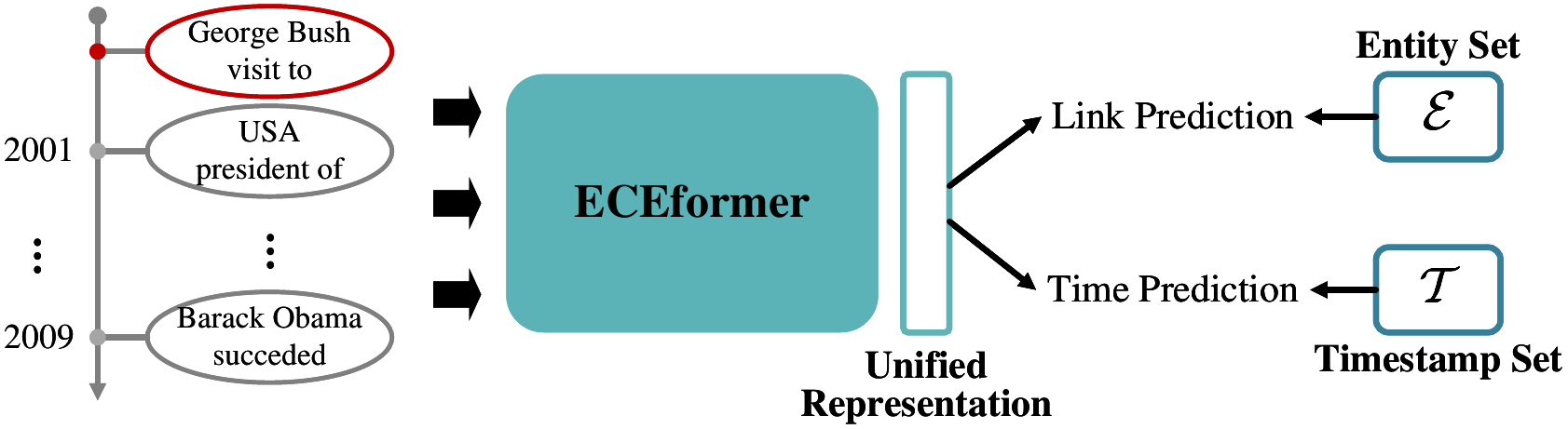} 
    \caption{}
    \label{fig:overall2}
\end{subfigure}
\caption{Illustration of the overall concept of our method. Part (a) displays a subgraph of TKG, which includes facts related to three U.S. Presidents. Part (b) illustrates our motivation for the ECEformer to learn a unified representation from an evolutionary chain of events, thereby simultaneously addressing link and time prediction tasks.}
\end{figure}
  
Knowledge graphs (KGs) store human-summarized prior knowledge in structured data formats and are widely applied in many downstream tasks, such as information retrieval \cite{SAFIR:agosti2020learning}, intelligent question answering \cite{BeamQA:atif2023beamqa}, and recommendation systems \cite{SimGCL:yu2022graph}. Traditional knowledge graphs employ triples ($s, p, o$) to record each fact or event in the knowledge base, where $s$ and $o$ respectively represent the subject and object entities, and $p$ denotes the logical predicate (or relation) connecting these two entities. e.g., \textit{(Barack Obama, presidentOf, USA)}. However, events in the real world often dynamically evolve over time. To extend the temporal nature of events, Temporal Knowledge Graphs (TKGs) incorporate temporal information into the traditional triples, resulting in quadruples ($s, p, o, \tau$). e.g., \textit{(Barack Obama, presidentOf, USA, 2009--2017)} and \textit{(Donald Trump, presidentOf, USA, 2017--2021)}.

Completing missing events for specific timestamps in TKGs via Temporal Knowledge Graph Reasoning (TKGR) has recently attracted growing attention from both academic and industrial communities. For example, the missing element \textit{Barack Obama} in the quadruple query \textit{(?, president of, USA, 2009--2017)} can be predicted via TKGR. Existing TKGR methods can be broadly categorized into two types according to the temporal scope of link prediction, i.e., interpolation and extrapolation \cite{ReNet:jin2020recurrent}. The former focuses on completing unknown knowledge in the past, whereas the latter concentrates on estimating unknown knowledge in the future. Concretely, interpolation reasoning methods aim to effectively integrate temporal information into the evolutionary trajectory of events. They typically utilize geographic and geometric information to model the relationships among elements within individual quadruples. Accordingly, HyTE \cite{HyTE:conf/emnlp/DasguptaRT18} employs a translation-based strategy \cite{TransE:conf/nips/BordesUGWY13} to measure the distance between entities and relations in the customized hyperplane, where each timestamp is associated with a corresponding hyperplane. Inspised by BoxE \cite{BoxE:conf/nips/AbboudCLS20}, BoxTE \cite{BoxTE:conf/aaai/MessnerAC22} additionally incorporates temporal information embedded through the relation-specific transition matrix within the foundational box-type relation modeling. TGeomE \cite{TGeomE:9713947}, a geometric algebra-based embedding approach, performs 4th-order tensor factorization and applies linear temporal regularization for time representation learning. These methods impose a rigid prior on KGs, relying solely on geometric modeling and geographic measurements. However, they ignore the implicit semantic information, failing to fully capture the complex graph structure and relational context. Therefore, it is necessary to \textbf{sufficiently explore the internal structure and semantic relationships within individual quadruples (Limitation I)}.

On the other hand, extrapolation reasoning methods aim to effectively leverage the query-associated historical information \cite{wang2023survey}. They generally utilize Graph Neural Network (GNN) models to capture structural characteristics within each snapshot and leverage Recurrent Neural Networks (RNNs) to explore the evolutionary characteristics across timestamps. Accordingly, RE-Net \cite{ReNet:jin2020recurrent} employs a multi-relational graph aggregator to capture the graph structure and an RNN-based encoder to capture temporal dependencies. RE-GCN \cite{REGCN:li2021temporal} leverages a relation-aware Graph Convolution Network (GCN) to capture the structural dependencies for the evolution unit and simultaneously captures the sequential patterns of all facts in parallel using gate recurrent components. RPC \cite{RPC:conf/sigir/0006MLLTWZL23} mines the information underlying the relational correlations and periodic patterns via two correspondence units: one is based on a relational GCN, and the other is based on Gated Recurrent Units (GRUs). However, this phased processing strategy can easily degrade the accuracy of predicting future events due to incorrect historical information. Therefore, it is necessary to adaptively correct the interference of incorrect historical information on prediction accuracy by \textbf{learning a unified representation of the contextual and temporal correlations among different quadruples (Limitation II)}.

Given a subgraph of TKG (as shown in Figure \ref{fig:overall1}), interpolation methods utilize geographical information to maximize structural differences between different quadruples. However, this type of approaches cannot effectively utilize semantic information in reasoning. Moreover, learning discriminative and powerful representations for each item within a quadruple is challenging, especially in modeling timestamps. For instance, differentiating between \textit{(Barack Obama, visitTo, France, 2009)} and \textit{(Donald Trump, visitTo, France, 2017)} necessitates precise characterizations of different names and timestamps. Although extrapolation methods consider neighborhood information, independently processing structural and temporal characteristics can lead to confusion in reasoning. For instance, under the timestamps of \textit{2009} and \textit{2017}, the structure of \textit{Barack Obama} and \textit{Donald Trump} with \textit{USA} and \textit{France} is consistent. Hence, the model can naturally infer the structure of \textit{George Bush} with \textit{USA} and \textit{France} in \textit{2001} should be consistent with the structure of \textit{Barack Obama} and \textit{Donald Trump}, which contradicts the facts. From this key observation, we believe that designing an end-to-end network for learning a unified representation rich in information from temporal subgraph inputs will enhance the performance of TKGR. This network adaptively extracts both intra-quadruple and inter-quadruple, together with temporal information.

According to the aforementioned analysis, we propose an innovative end-to-end Transformer-based network to learn the evolutionary chain of events (dubbed ECEformer) for TKGR. Figure \ref{fig:overall2} illustrates the motivation of our method. To easily illustrate the structural and contextual relationship between the query node and the adjacent nodes, we form an evolutionary chain of events (ECE) by extracting query-specific neighbors from the subgraph. Subsequently, our proposed Transformer-based model learns the unified representation for the ECE via two novel modules, namely, ECE Representation Learning (ECER) and Mixed-Context Knowledge Reasoning (MCKR). Specifically, to overcome \textbf{Limitation I}, ECER employs a Transformer-based encoder to embed each event in the ECE, such as \textit{(USA, president of, 2001)} corresponding to a specific query \textit{George Bush}. To overcome \textbf{Limitation II}, MCKR induces the embeddings of each event and enhances interaction within and between quadruples via crafting an MLP-based information mixing layer. In addition, to enhance the timeliness of the events, we devise an additional time prediction task to imbue effective temporal information within the learned unified representation.

In summary, our main contributions are three-fold:
\begin{itemize}
\item We propose an innovative temporal knowledge graph reasoning method, namely ECEformer. Experimental results verify the state-of-the-art performance of our method on six publicly available datasets.
\item Proposing a Transformer-based encoder for ECE representation learning (ECER) aims to explore the internal structure and semantic relationships within individual quadruples.
\item Proposing an MLP-based layer for Mixed-Context Knowledge Reasoning (MCKR) aims to learn a unified representation of the contextual and temporal correlations among different quadruples.
\end{itemize}

\section{Related Work}
\subsection{Interpolation-based TKGR}
Interpolation-based TKGR methods estimate missing facts or events by identifying consistent trends within a TKG. Based on the process strategy for temporal information, we categorize them into two types: timestamp-independent and timestamp-specific methods.

Timestamp-independent methods treat timestamps as independent units, equivalent to entities or relations, and do not apply additional operations on the timestamps. Generally, these methods \cite{TTransE:conf/www/LeblayC18,TComplex:conf/iclr/LacroixOU20,ChronoR:conf/aaai/SadeghianACW21,BoxTE:conf/aaai/MessnerAC22} directly associate timestamps to the corresponding entities and relations based on the foundation of static KG models. For example, Leblay \textit{et al.} \cite{TTransE:conf/www/LeblayC18} extended the classic TransE \cite{TransE:conf/nips/BordesUGWY13} model to TKGR by concatenating timestamp embeddings with relation embeddings. Inspired by BoxE \cite{BoxE:conf/nips/AbboudCLS20}, Messner \textit{et al.} \cite{BoxTE:conf/aaai/MessnerAC22} directly introduced timestamp embeddings on its basis. However, timestamp-independent methods are usually limited in capturing the temporal information of evolutionary events. Timestamp-specific methods embed the temporal information and learn the evolution of entities and relations via timestamp-specific functions. To effectively utilize the structural and semantic information of the time, they ingeniously designed various time-specific functions, such as diachronic embedding functions \cite{DE-SimplE:conf/aaai/GoelKBP20}, time-rotating functions \cite{TeRo:conf/coling/XuNAYL20,RotateQVS:conf/acl/ChenWLL22}, time-hyperplane functions \cite{HyTE:conf/emnlp/DasguptaRT18,TGeomE:9713947,TeLM:xu2021temporal,HyIE:zhang2023hybrid}, and non-linear embedding functions \cite{ATiSE,DyERNIE:conf/emnlp/HanCMT20,T-GAP:jung2021learning,SANe:li2022each,QDN:wang2023qdn}. Specifically, Goel \textit{et al.} \cite{DE-SimplE:conf/aaai/GoelKBP20} proposed a diachronic entity embedding function to characterize entities at any timestamp. Chen \textit{et al.} \cite{RotateQVS:conf/acl/ChenWLL22} defined the spatial rotation operation of entities around the time axis. Zhang \textit{et al.} \cite{HyIE:zhang2023hybrid} learned spatial structures interactively between the Euclidean, hyperbolic and hyper-spherical spaces. Han \textit{et al.} \cite{DyERNIE:conf/emnlp/HanCMT20} explored evolving entity representations on a mixed curvature manifold using a velocity vector defined in the tangent space at each timestamp. Despite the great efforts made by these studies, they are limited in predicting future events \cite{DREAM:conf/sigir/ZhengYCNCZ23}.

\subsection{Extrapolation-based TKGR}
Extrapolation-based TKGR methods predict future facts or events by learning effective embeddings from historical snapshots. As knowledge subgraphs at specific timestamps, historical snapshots inherently contain rich structural and semantic information. Extrapolation methods typically employ deep learning techniques such as Graph Neural Networks (GNNs) \cite{REGCN:li2021temporal,HGLS:zhang2023learning,L2TKG:zhang2023learning}, Recurrent Neural Networks (RNNs) \cite{ReNet:jin2020recurrent,TiRGN:li2022tirgn}, and Reinforcement Learning (RL) \cite{TITer:sun2021timetraveler,DREAM:conf/sigir/ZhengYCNCZ23} to extract features from historical snapshots.

To elaborate further, GNN-based methods are naturally applied to TKGR due to the inherent topological structure of historical snapshots. Schlichtkrull \textit{et al.} \cite{R-GCN:conf/esws/SchlichtkrullKB18} early utilized Graph Convolutional Networks (GCN) to model the multi-relational features of traditional KG. To incorporate the temporal characteristics in TKGs, Jin \textit{et al.} \cite{ReNet:jin2020recurrent} additionally introduced an RNN-based sub-network specifically for recursively encoding past facts. Li \textit{et al.} \cite{REGCN:li2021temporal} presented the recurrent evolution network based on GCN, which learns the evolutionary representations of entities and relations at each timestamp by modeling the KG sequence recurrently. To handle the large-scale inputs and non-independent and identically distributed data, Deng \textit{et al.} \cite{deng2020dynamic} combined GCN and GRU for simultaneously predicting concurrent events of multiple types and inferring multiple candidate participants. RL-based methods are becoming popular to improve the interpretability of reasoning methods, especially in question-answering tasks. By mimicking the human mechanism of searching for historical information and meticulous reasoning, CluSTeR \cite{li2021search} implements an interpretable prediction for TKGR using a two-stage approach of clue searching and temporal reasoning. DREAM \cite{DREAM:conf/sigir/ZhengYCNCZ23} introduces an adaptive reinforcement learning model based on attention mechanism, addressing the challenges in traditional RL-based TKG reasoning, specifically the lack of capturing temporal evolution and semantic dependence jointly and the over-reliance on manually designed rewards. Furthermore, some other types of methods also achieve good performance in TKGR. For example, TLogic \cite{TLogic:liu2022tlogic} introduces an explainable framework based on temporal logical rules, leveraging temporal random walks to provide explanations while maintaining time consistency. Faced with the challenge of predicting future facts for newly emerging entities based on extremely limited observational data, MetaTKGR \cite{wang2022learning} dynamically adjusts its strategies for sampling and aggregating neighbors from recent facts and introduces a temporal adaptation regularizer to stabilize the meta-temporal reasoning process over time. In summary, extrapolation methods generally process structural and temporal characteristics independently. This approach ignores the intrinsic correlation between structural and temporal information, limiting the capability to extract a unified representation of both features.

\subsection{Transformer in Knowledge Graph}
With the Transformer achieving excellent performance in numerous tasks in natural language processing and computer vision, it has been introduced into knowledge graph tasks. Leveraging the powerful learning capability of Transformers for sequential data structures and contexts, numerous Transformer-based methods \cite{KG-BERT:yao2019kg,HittER:chen2021hitter,MKGformer:chen2022hybrid,hu2023hyperformer} have emerged for traditional KGs. For example, Chen \textit{et al.} \cite{HittER:chen2021hitter} proposed two different Transformer blocks for hierarchically learning representations of entities and relations. In addition, some methods also achieve remarkable results in downstream tasks of KGs. Specifically, Hu \textit{et al.} \cite{hu-etal-2022-transformer} utilized a Transformer-based framework for encoding the content of neighbors of an entity for entity typing task. Meanwhile, Xu \textit{et al.} \cite{xu2022ruleformer} applied it for context-aware rule mining over KG. To serve diverse KG-related tasks, Zhang \textit{et al.} \cite{KGTransfo:zhang2023structure} proposed a uniform knowledge representation and fusion module via structure pretraining and prompt tuning.

Transformer enables the exploration of structural associations within each historical snapshot and captures the temporal relationships among different historical snapshots to accomplish link prediction tasks \cite{wang2023survey}. GHT \cite{GHT:sun2022graph} involves two variants of Transformer and a relational continuous-time encoding function, aiming to mitigate the interference of facts irrelevant to the query and enhancing the capability of the model to learn long-term evolution. To effectively embed the rich information in the query-specific subgraph, SToKE \cite{SToKE:conf/acl/GaoHKHQ023} learns joint structural and temporal contextualized knowledge embeddings employing the pre-trained language model. Overview: While the Transformer framework has wide application in traditional KGs, its exploration in the field of TKGR remains relatively limited.

\begin{figure*}[t]
\centering
\includegraphics[width=1.0\linewidth]{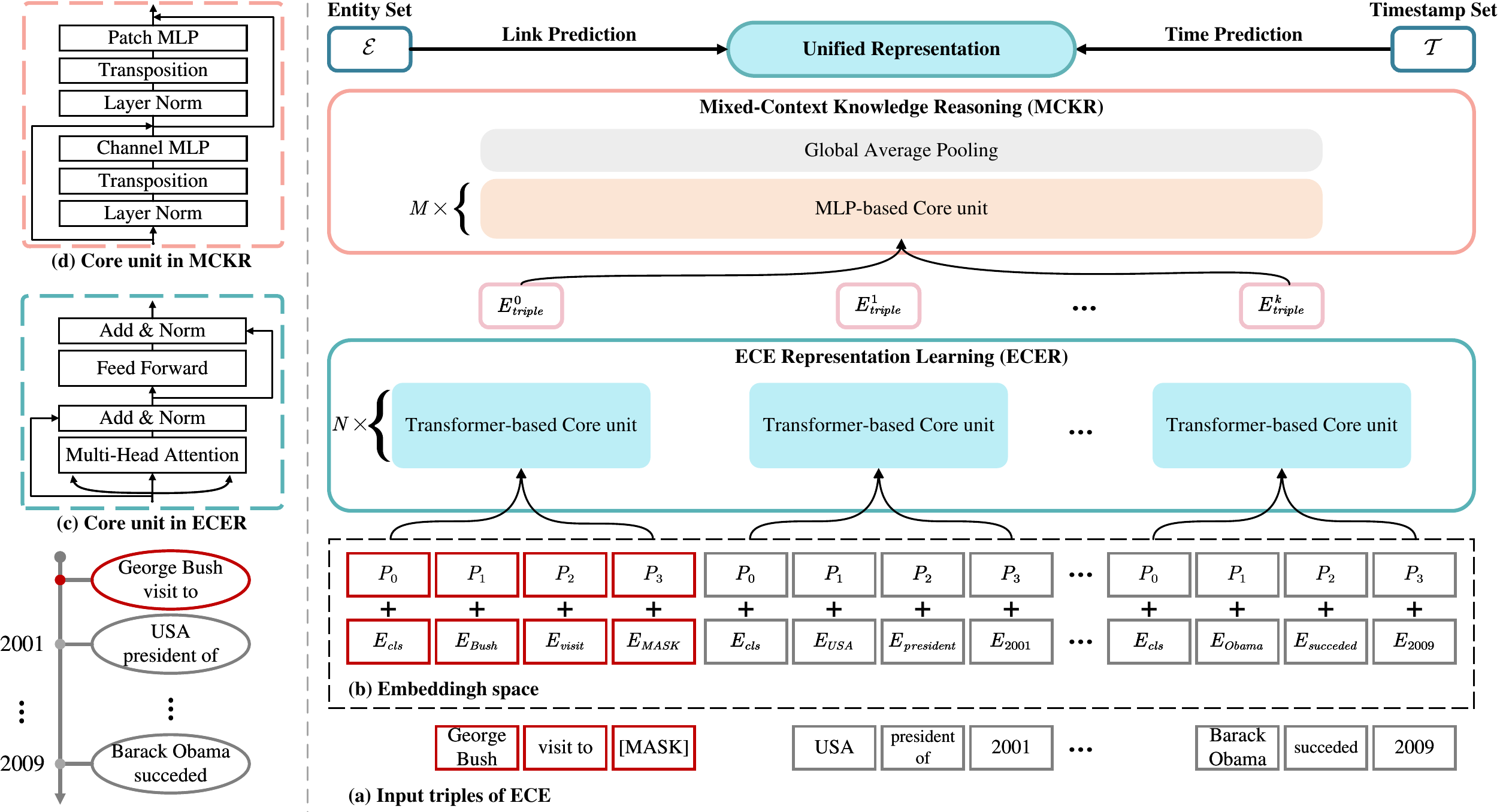}
\caption{The architecture of the proposed ECEformer. Given an evolutionary chain of events, ECEformer obtains input sequences (detailed in (a)) from each branch of the chain and converts each token into a concatenation of semantic embeddings $E$ and positional embeddings $P$ (detailed in (b)). Subsequently, these branch embeddings $E_{triple}$ are derived by the ECER, which is based on the Transformer encoder (detailed in (c)). The contextual information from different branch embeddings is then inductively processed by the MLP-based MCKR (detailed in (d)), culminating in a unified representation.}
\label{fig:arch}
\end{figure*}

\section{Methodology}
\subsection{Preliminary}
Note that TKG $\mathcal{G} =(\mathcal{E}, \mathcal{R}, \mathcal{T}, \mathcal{Q})$ denotes a directed multi-relation graph with timestamped edges between entities, where $\mathcal{E}\in\mathbb{R}^{|\mathcal{E}|\times d}$, $\mathcal{R}\in\mathbb{R}^{|\mathcal{R}|\times d}$, and $\mathcal{T}\in\mathbb{R}^{|\mathcal{T}|\times d}$ are the set of entities, relations and timestamps, respectively. $\mathcal{Q} = \{ (s, p, o, \tau)\ |\ s,$ $ o \in\mathcal{E}, p \in\mathcal{R}, \tau \in\mathcal{T}\}$ is a set of quadruples in $\mathcal{G}$, where $s$, $o$, $p$ and $\tau$ are a subject entity, an object entity, the relation between them, and a timestamp, respectively. For the link prediction task, we infer the missing object entity by converting $(s, p, ?, \tau)$. Similarly, we infer the missing subject entity by converting $(o, p^{-1}, ?, \tau)$, where we employ a reciprocal predicate $p^{-1}$ to distinguish this case.

\textbf{Evolutionary Chain of Events.} 
The evolutionary chain of events (ECE) transforms structural and temporal contexts related to the query-specific neighborhood subgraph into a structured knowledge sequence. Specifically, for the query $(s,p,?,\tau)$, ECE is conceptualized as a singly linked list $\mathcal{C}=\{C_0, C_1, \cdots, C_k\}$, comprising a chronologically arranged collection of events relevant to the subject $s$ of the query. Notably, each $C$ represents an event related to $s$, depicted by a triplet $(e, p, \tau)$, where $C_0$ signifies the query and $C_k$ denotes adjacent facts, $k$ being the maximum adjacency count, $e$ can serve as either the subject entity or the object entity. For example, for the query \textit{(George Bush, visitTo, ?, 2002)} as shown in Figure \ref{fig:overall1}, the ECE can be formulated as: $\mathcal{C}=$\textit{\{(George Bush, visitTo, 2002), (USA, presidentOf, 2001), (Barack Obama, succeded, 2009)\}}. The formal expression is illustrated in Figure \ref{fig:overall2}. Hence, Our ECE can structurally represent both intra-quadruples and inter-quadruples. By encapsulating the dynamic evolution of entities and their interrelations within the event chain, the ECE can capture the historical and current events related to the subject entity.

\subsection{Overview of ECEformer}
As an innovative end-to-end Transformer-based network, ECEformer learns the evolutionary chain of events (ECE) for TKGR. This work defines the subgraph composed of adjacent nodes that evolve over time along with the query entity as the evolutionary chain of events (ECE). Different from existing Transformer-based models, ECEformer crafts specific modules for handling intra-quadruples and inter-quadruples, respectively. As for intra-quadruples, ECE representation learning (ECER) based on a Transformer encoder is designed to explore the internal structure and semantic relationship within individual quadruples from the input ECE. On the other hand, as for the inter-quadruples, mixed-context knowledge reasoning (MCKR) involves a novel information mixing layer, and the iterative transposition mechanism is designed to explore the contextual and temporal correlations among different quadruples. The architecture of ECEformer is shown in Figure \ref{fig:arch}, and more details will be illustrated in the following sections.

\subsection{Learning Representation for Evolutionary Chain of Events}
The first key to effectively learning ECE lies in the comprehensive extraction and analysis of each triplet corresponding to the sub-nodes on the ECE. Every constituent node of the ECE stores a historical event associated with the central entity, necessitating an investigation into its inherent structural and temporal characteristics. Inspired by the effectiveness of the Transformer architecture in learning inter-data dependencies, we introduce an ECER module utilizing a Transformer encoder that distills the embedding of each triplet. The details are shown in Figure \ref{fig:arch}c. Specifically, ECER consists of $N$ core units, each consisting of alternating layers of multi-head attention (MHA), layer normalization (LN), and MLP-based feed-forward (FF). Among them, both MHA and FF are followed by a LN accompanied by the residual connection (RC). The FF contains two layers with the GELU non-linear activation function. This process can be formulated as:
\begin{equation}
y=l\circ r\circ f\circ l\circ r\circ h(x;W_h,W_l,W_f),
\end{equation}
where $h,l$ and $f$ denote MHA function with weight $W_h$, LN function with weight $W_l$, and FF function with weight $W_f$, respectively. $r$ denotes the RC function, which performs an operation that directly combines the input with the output of the intermediate layer. $x$ denotes the input sequence. $y$ denotes the output embedding. $\circ$ denotes the composite function operator. Note that the MHA relies on the scaled dot-product attention mechanism to learn semantic interactions between the current token and all tokens within the sequence, thereby thoroughly exploring the structural and semantic information within individual events. Additionally, by directly inputting timestamps as tokens into ECER, the model can semantically extract temporal information within individual events.

Given a query $q(s,p,?,\tau)$ and its ECE $\mathcal{C}_q$, we treat each component within the branch of ECE as an individual token, sequentially unfolding the chain structure based on temporal order. This process yields a new input, as illustrated in Figure \ref{fig:arch}a. For each triplet on every branch of the ECE, we append a special token, i.e., $[CLS]$. Subsequently, we employ the concatenation of semantic embeddings $E$ and positional embeddings $P$ in the embedding space to represent each input token, as illustrated in Figure \ref{fig:arch}b. This process can be formulated as:
\begin{equation}
E_{inp}^{i}=E_i+P_j,\left\{i\in\mathcal{E}\cup\mathcal{R}\cup\mathcal{T},j\in[0,1,2,3]\right\},
\end{equation}
where $E_{inp}^{i}$ denotes $i-th$ token in the input sequence, $E_i$ denotes the feature embedding of $i-th$ token, and $P_j$ denotes the position embedding. These embeddings are iteratively fed into the ECER module $\mathcal{F}_{ECER}:\mathbb{R}^{4\times d}\rightarrow\mathbb{R}^d$. Eventually, the output consists of intermediate embeddings $E_{triple}^i,i=0,1,\cdots,k$, which represent the structural and semantic information implied in the triples corresponding to each branch.

\subsection{Mixed-Context Knowledge Reasoning}
After encoding through the ECER, the embedded representations have effectively abstracted the relevant information of individual events. We introduce the query-specific contextual information to further enhance the accuracy of reasoning. Inspired by \cite{tolstikhin2021mlp}, we introduce the MLP-based MCKR module to learn the multi-dimensional information interaction of the inputs by leveraging two mixing strategies. The details are illustrated in Figure \ref{fig:arch}d.

Different from the single event input in ECER, the MCKR module processes the sequence of $1+k$ embedding representations of ECE. This forms a two-dimensional real-valued input matrix $\mathcal{M}\in\mathbb{R}^{(1+k)\times d}$. The MCKR consists of $M$ core units, each consisting of two key mixing layers: the channel MLP and the patch MLP. Additionally, each mixing unit is preceded by a layer normalization and a transposition layer. The channel MLP operates on the columns of $\mathcal{M}$, mapping each column from $\mathbb{R}^{1+k}$ to $\mathbb{R}^{1+k}$. It aims to facilitate information interaction across different event embeddings within the same feature dimension. Conversely, the patch MLP operates on the rows of $\mathcal{M}$, mapping each row from $\mathbb{R}^{d}$ to $\mathbb{R}^{d}$. It aims to facilitate information interaction within the same event embedding across different feature dimensions. This process can be formulated as:
\begin{equation}
\dot{\mathcal{M}}_{*,i}=M_{*,i}+W_2\sigma(W_1\textnormal{Norm}(\mathcal{M})_{*,i}), \textnormal{ for } i=1,2,\cdots,d
\end{equation}
\begin{equation}
\ddot{\mathcal{M}}_{j,*}=\dot{\mathcal{M}}_{j,*}+W_4\sigma(W_3\textnormal{Norm}(\dot{\mathcal{M}})_{j,*}), \textnormal{ for } j=1,2,\cdots,k+1
\end{equation}
where $\sigma$ is the non-linear activation function (GELU). Each MLP block in MCKR contains two fully-connected layers, $W_1$ and $W_2$ are hidden weights in the channel MLP, $W_3$ and $W_4$ are hidden weights in the patch MLP. 

Following the methodology outlined in \cite{tolstikhin2021mlp}, our MCKR employs the same parameter to translate every column (row) for channel MLP (patch MLP). This parameter sharing mechanism endows the network with positional invariance, enhancing its ability to proficiently process sequential data. Moreover, this approach also circumvents the substantial increase in model complexity that typically accompanies dimensional augmentation. Subsequently, we leverage a global average pooling layer to distill the mixed-context $\ddot{M}$, which can be formulated as:
\begin{equation}
\mathcal{U}_{ECE}=\textnormal{GlobalAvgPool}(\ddot{M}),
\end{equation}
where $\mathcal{U}_{ECE}$ denotes the unified representation of ECE. Given a quadruple with either the subject or the object missing, the task of link prediction can be accomplished by computing the similarity between $\mathcal{U}_{ECE}$ and all candidate entities. Link prediction can be mathematically formulated as:
\begin{equation}
p(e_{gt}|q)=Softmax(Sim(\mathcal{U}_{ECE},\mathcal{E})),
\end{equation}
where $Sim$ denotes the similarity function between $\mathcal{U}_{ECE}$ and a candidate entity, $q$ denotes the predicted result, obtained by selecting the $top-1$ entity based on similarity scores. $p(e_{gt}|q)$ denotes the occurrence probability that $q$ is the target entity $e_{gt}$.

\begin{table}[t]
\caption{Statistics of the Experimental Datasets.}
\label{tab:dataset}
\scalebox{0.8}{
\begin{tabular}{*{8}c}
\toprule
Dataset & |$\mathcal{E}$| & |$\mathcal{R}$| & |$\mathcal{T}$| & \textit{Time} & $N_{train}$ & $N_{valid}$ & $N_{test}$ \\
\midrule
GDELT & 500 & 20 & 366 & 15 mins & 2,735,685 & 341,961 & 341,961 \\
ICEWS05-15 & 10,488 & 251 & 4,017 & 24 hours & 386,962 & 46,092 & 46,2775 \\
ICEWS18 & 23,033 & 256 & 7,272 & 24 hours & 373,018 & 45,995 & 49,545 \\
ICEWS14 & 7,128 & 230 & 365 & 24 hours & 63,685 & 13,823 & 13,222 \\
YAGO11K & 10,623 & 10 & 73 & 1 year & 16,406 & 2,050 & 2,051 \\
Wikidata12K & 12,554 & 24 & 84 & 1 year & 32,497 & 4,062 & 4,062 \\
\bottomrule
\end{tabular}}
\end{table}

\begin{table*}[t]
\caption{Performance of link prediction on GDELT, ICEWS05-15, and ICEWS18. The best results are marked in bold, the second results are marked by \underline{underlining}.}
\label{tab:sota1}
\scalebox{1.0}{
\begin{tabular}{*{13}c}
\toprule
\multirow{2}{*}{Model} & \multicolumn{4}{c}{GDELT} & \multicolumn{4}{c}{ICEWS05-15} & \multicolumn{4}{c}{ICEWS18} \\
& MRR & Hits@1 & Hits@3 & Hits@10 & MRR & Hits@1 & Hits@3 & Hits@10 & MRR & Hits@1 & Hits@3 & Hits@10 \\
\midrule
TITer & 18.19 & 11.52 & 19.20 & 31.00 & 47.60 & 38.29 & 52.74 & 64.86 & 29.98 & 22.05 & 33.46 & 44.83 \\
TLogic & 19.80 & 12.20 & 21.70 & 35.60 & 46.97 & 36.21 & 53.13 & 67.43 & 29.82 & 20.54 & 33.95 & 48.53 \\
DREAM & 28.10 & 19.30 & 31.10 & 44.70 & 56.80 & 47.30 & 65.10 & 78.60 & \underline{39.10} & \underline{28.00} & \underline{45.20} & \bf{62.70} \\
\midrule
RE-GCN & 19.69 & 12.46 & 20.93 & 33.81 & 48.03 & 37.33 & 53.90 & 68.51 & 32.62 & 22.39 & 36.79 & 52.68 \\
TiRGN & 21.67 & 13.63 & 23.27 & 37.60 & 49.84 & 39.07 & 55.75 & 70.11 & 33.58 & 23.10 & 37.90 & 54.20 \\
HGLS & 19.04 & 11.79 & - & 33.23 & 46.21 & 35.32 & - & 67.12 & 29.32 & 19.21 & - & 49.83 \\
$L^2$TKG & 20.53 & 12.89 & - & 35.83 & 57.43 & 41.86 & - & 80.69 & 33.36 & 22.15 & - & 55.04 \\
RPC & 22.41 & 14.42 & 24.36 & 38.33 & 51.14 & 39.47 & 57.11 & 71.75 & 34.91 & 24.34 & 38.74 & \underline{55.89} \\
\midrule
GHT & 20.04 & 12.68 & 21.37 & 34.42 & 41.50 & 30.79 & 46.85 & 62.73 & 27.40 & 18.08 & 30.76 & 45.76 \\
SToKE & \underline{37.10} & \underline{29.00} & \underline{39.90} & \underline{52.50} & \underline{71.20} & \underline{60.50} & \underline{79.00} & \bf{88.50} & - & - & - & - \\
\midrule
ECEformer (Ours) & \bf{51.19} & \bf{38.72} & \bf{59.41} & \bf{71.10} & \bf{77.29} & \bf{73.11} & \bf{79.39} & \underline{85.17} & \bf{44.80} & \bf{39.18} & \bf{46.89} & {55.31} \\
\bottomrule
\end{tabular}}
\end{table*}

\subsection{Enhancement of Temporal Information}
Considering that events in TKGs often coincide with specific timestamps, the unified representations learned from link prediction might be trivial solutions facilitated by contextual information. Such representations  will introduce spurious correlations, because they only consider entities and relations while neglecting the temporal aspect of events. To address this issue, we propose a marked time prediction task to enhance the exploration of temporal information during the contextualization process. Specifically, we utilize a special token $[MASK]$ to replace the timestamp $\tau$ in the query-specific branch of ECE. In other words, we transform the $C_0(s,p,\tau)$ in ECE to $C_0(s,p,[MASK])$. This direct masking strategy introduces perturbations to the original timestamps, thereby prompting the model to learn contextual information. To promote the model's capability of understanding temporal information, we train the model using masked inputs to recover the perturbed timestamp. This design can encourage the model to assimilate contextual information and deduce the occurrence time of the query-specific event. The methodology is analogous to link prediction, where we directly utilize the unified representation $\mathcal{U}_{ECE}$ to predict the correct timestamp. The prediction probability $p(\tau_{gt}|q)$ can be formulated as:
\begin{equation}
p(\tau_{gt}|q)=Softmax(Sim(\mathcal{U}_{ECE},\mathcal{T})),
\end{equation}
where $p(\tau_{gt}|q)$ denotes the probability that predicted timestamp $q$ corresponds to the target timestamp $\tau_{gt}$.

\subsection{Training and Optimization}
For link prediction and time prediction tasks, we employ the cross-entropy function to calculate the loss during the training process. The loss function can be formulated as:
\begin{equation}
\mathcal{L}=-\sum_{(s,p,o,\tau)\in\mathcal{G}}\log(p(e_{gt}|q)) + \lambda\log(p(\tau_{gt}|q)),
\end{equation}
where $(s,p,o,\tau)\in\mathcal{G}$ represents the historical events in the training set, $\lambda$ weights the time prediction task.

Additionally, for diversifying our training data and reducing memory cost, we implement an ECE sampling strategy similar to the edge dropout regularization \cite{HittER:chen2021hitter}. This approach samples only a portion of the neighborhood information of the query subject and filters out the ground truth target entity from the sampled neighborhood information.

\begin{table*}[t]
\caption{Performance of link prediction task on ICEWS14, YAGO11K, and Wikidata12K. The best results are marked in bold, the second results are marked by \underline{underlining}.}
\label{tab:sota2}
\scalebox{1.0}{
\begin{tabular}{*{13}c}
\toprule
\multirow{2}{*}{Model} & \multicolumn{4}{c}{ICEWS14} & \multicolumn{4}{c}{YAGO11K} & \multicolumn{4}{c}{Wikidata12K} \\
& MRR & Hits@1 & Hits@3 & Hits@10 & MRR & Hits@1 & Hits@3 & Hits@10 & MRR & Hits@1 & Hits@3 & Hits@10 \\
\midrule
TimePlex & 60.40 & 51.50 & - & 77.11 & 23.64 & 16.92 & - & 36.71 & 33.35 & 22.78 & - & 53.20 \\
TeLM & 62.50 & 54.50 & 67.30 & 77.40 & 19.10 & 12.90 & 19.40 & 32.10 & 33.20 & 23.10 & 36.00 & 54.20 \\
TeRo & 56.20 & 46.80 & 62.10 & 73.20 & 18.70 & 12.10 & 19.70 & 31.90 & 29.90 & 19.80 & 32.90 & 50.70 \\
RotateQVS & 59.10 & 50.70 & 64.20 & 75.40 & 18.90 & 12.40 & 19.90 & 32.30 & - & - & - & - \\
SANe & 63.80 & 55.80 & 68.80 & 78.20 & 25.00 & 18.00 & \underline{26.60} & \bf{40.10} & \underline{43.20} & \underline{33.10} & 48.30 & \underline{64.00} \\
NeuSTIP & - & - & - & - & \underline{25.23} & \underline{18.45} & - & {37.76} & 34.78 & 24.38 & - & 53.75 \\ 
QDN & 64.30 & 56.70 & 68.80 & 78.40 & 19.80 & 13.10 & 20.10 & 32.80 & - & - & - & - \\
LCGE & \underline{66.70} & \underline{58.80} & \underline{71.40} & \bf{81.50} & - & - & - & - & 42.90 & 30.40 & \underline{49.50} & \bf{67.70} \\
TGeomE & 62.90 & 54.60 & 68.00 & 78.00 & 19.50 & 13.00 & 19.60 & 32.60 & 33.30 & 23.20 & 36.20 & 54.60 \\
HyIE & 63.10 & 56.30 & 68.70 & 78.60 & 19.10 & 12.50 & 20.10 & 32.60 & 30.10 & 19.70 & 32.80 & 50.60 \\
\midrule
ECEformer (Ours) & \bf{71.70} & \bf{67.31} & \bf{73.60} & \underline{80.30} & \bf{25.63} & \bf{19.48} & \bf{26.88} & \underline{37.84} & \bf{47.81} & \bf{41.35} & \bf{49.96} & {60.35} \\
\bottomrule
\end{tabular}}
\end{table*}

\section{Experiments}
In this section, the experimental settings are first introduced from four aspects, including datasets, evaluation metrics, compared baselines, and implementation details. Then, we comprehensively analyze the proposed ECEformer also from three aspects, i.e., \textbf{superiority}, \textbf{effectiveness}, and \textbf{sensitivity}.

\subsection{Experimental Setup}
\subsubsection{Datasets and Evaluation Metrics.} We evaluate the reasoning performance of our ECEformer on six TKG benchmark datasets, including GDELT, ICEWS05-15, ICEWS18, ICEWS14, YAGO11K, and Wikidata12K. According to \cite{RPC:conf/sigir/0006MLLTWZL23}, we also split the datasets of GDELT and ICEWS05-15/18/14 into train/valid/test by timestamps. According to \cite{HyTE:conf/emnlp/DasguptaRT18}, we split  the datasets of YAGO11K and Wikidata12K into train/valid/test by time intervals. The statistical information of datasets is shown in Table \ref{tab:dataset}, in which \textit{Time} represents time granularity. Two widely used evaluation metrics are adopted to quantify the performance, namely $MRR$ and $Hits@k$. $MRR$ represents the mean reciprocal rank of the inferred true entity among the queried candidates, and $Hits@k$ measures the proportion of instances where the true entity appears among the top $k$ ranked candidates.

\subsubsection{Compared Baselines.}
Considering that different methods are validated on various datasets, we divided our experiments into two groups according to dataset size for a fair comparison. We compared with twenty state-of-the-art (SOTA) models. Specifically, the extrapolation baselines for GDELT, ICEWS05-15, and ICEWS18 consist of RL-based models (TITer \cite{TITer:sun2021timetraveler}, TLogic \cite{TLogic:liu2022tlogic}, and DREAM \cite{DREAM:conf/sigir/ZhengYCNCZ23}), GNN-based models (RE-GCN \cite{REGCN:li2021temporal}, TiRGN \cite{TiRGN:li2022tirgn}, HGLS \cite{HGLS:zhang2023learning}, $L^2$TKG \cite{L2TKG:zhang2023learning}, and RPC \cite{RPC:conf/sigir/0006MLLTWZL23}), and Transformer-based models (GHT \cite{GHT:sun2022graph} and SToKE \cite{SToKE:conf/acl/GaoHKHQ023}). The interpolation baselines chosen for smaller datasets (ICEWS14, YAGO11K, and Wikidata12K) include TimePlex \cite{TimePlex:jain2020temporal}, TeLM \cite{TeLM:xu2021temporal}, TeRo \cite{TeRo:conf/coling/XuNAYL20}, RotateQVS \cite{RotateQVS:conf/acl/ChenWLL22}, SANe \cite{SANe:li2022each},  NeuSTIP \cite{NeuSTIP:singh2023neustip}, QDN \cite{QDN:wang2023qdn}, LCGE \cite{LCGE:niu2023logic}, TGeomE \cite{TGeomE:9713947}, and HyIE \cite{HyIE:zhang2023hybrid}.

\subsubsection{Implementation Details.}
All experiments are conducted on a single NVIDIA RTX A6000. The implementation\footnote{https://github.com/seeyourmind/TKGElib} of our model is built upon the open-source framework LibKGE\footnote{https://github.com/uma-pi1/kge}. The dimension of the input embedding is set to 320. The hidden widths of the feed-forward layer inside ECER and the channel (patch) MLP inside MCKR are set to 1024. The maximum neighborhood of ECE is set to 50. We configure the core unit numbers for ECER and MCKR to be $N=3$ and $M=6$, respectively. The activation function in our network is GELU. 

We train our model via the Adamax \cite{Adam:journals/corr/KingmaB14} with an initial learning rate of 0.01 and an L2 weight decay rate of 0.01. We employed the warm-up training strategy, wherein the learning rate commenced from zero and increased linearly during the initial 10\% of the training steps, followed by a linear decrease throughout the remaining training process. We train the model with a batch size of 512 for a maximum of 300 epochs, employing an early stopping mechanism based on MRR in the validation set. Detailed discussions on the effects of the weight coefficient $\lambda$, the number of core units $N$ and $M$ are in the subsection \textbf{Sensitivity Analysis}.

\subsection{Performance Comparison for Superiority}
We compare our model with 20 SOTA models ranging from 2020 to 2023 on six benchmark datasets. We report the result in Table \ref{tab:sota1} and Table \ref{tab:sota2}, where the best performances are highlighted in bold, while the second-best are marked by underlining. A noteworthy observation is that our ECEformer significantly outperforms other baseline models across all datasets in terms of $MRR$ and $Hits@k$. Specifically, ECEformer achieves performance improvements over the second-best models by 14.09\%, 6.09\%, 5.70\%, 5.00\%, 0.40\%, and 4.61\% on $MRR$, respectively. Additionally, it achieves gains of 9.72\%, 12.61\%, 11.18\%, 8.51\%, 1.03\%, and 8.25\% on $Hits@1$, respectively. This significant margin of improvement demonstrates the superior performance of our ECEformer.

In particular, our model exhibits notable performance in four metrics on GDELT: $MRR$, $Hits@1$, $Hits@3$, and $Hits@10$, achieving respective scores of 51.19\%, 38.72\%, 59.41\%, and 71.10\%. Compared with RL-based methods, our ECEformer outperforms DREAM by 23.09\% in $MRR$, by 9.42\% in $Hits@1$, by 28.31\% in $Hits@3$, and by 26.4\% in $Hits@10$, respectively. Compared with GNN-based methods, our ECEformer outperforms RPC by 28.78\% in $MRR$, by 24.30\% in $Hits@1$, by 35.05\% in $Hits@3$, and by 32.77\% in $Hits@10$, respectively. Compared with Transformer-based methods, our ECEformer outperforms SToKE by 14.09\% in $MRR$, by 9.72\% in $Hits@1$, by 19.51\% in $Hits@3$, and by 18.6\% in $Hits@10$, respectively. This indicates that our model can effectively learn the unified representations from frequently changing historical events in datasets characterized by fine-grained temporal granularity. Based on the comparative results from the ICEWS (i.e., ICEWS05-15/18/14), ECEformer consistently surpasses all the baselines on the $MRR$, $Hits@1$, and $Hits@3$ metrics. However, it lags behind the second-best model in $Hits@10$. This situation also occurs in the YAGO11K and Wikidata12K. This suggests that although our model demonstrates high precision in predictions, there is still room for improvement in its recall capability. Besides, compared to other datasets, our method shows a relatively slight performance improvement on YAGO11K. This is attributed to 1) the issue of missing timestamps in the YAGO11K dataset, and 2) the NeuSTIP method additionally introduces Allen algebra \cite{NeuSTIP:singh2023neustip} to refine temporal rules and designs a specialized time prediction evaluation function to capture temporal information. Nevertheless, the performance gain achieved by our method indicates that the proposed masked time prediction task can mitigate challenges arising from low data quality.

\begin{table}[t]
\caption{Ablation study on different module combinations on ICEWS14 and Wikidata12K. $
TP$ represents the time prediction task. The best results are marked in bold.}
\label{tab:abl}
\scalebox{1.0}{
\begin{tabular*}{\linewidth}{@{\extracolsep{\fill}}*{5}{c}}
\toprule
\multirow{2}{*}{Model} & \multicolumn{2}{c}{ICEWS14} & \multicolumn{2}{c}{Wikidata12K} \\
& MRR & Hits@10 & MRR & Hits@10 \\
\midrule 
ECER & 70.60 & \bf{80.40} & 43.49 & 57.08\\ 
ECEformer (w/o TP) & 69.4 & 79.1 & 46.12 & 58.54\\
ECEformer & \bf{71.70} & 80.30 & \bf{47.81} & \bf{60.35}\\
\bottomrule
\end{tabular*}}
\end{table}

\subsection{Ablation Study for Effectiveness}
To analyze the contribution of ECEformer components, we validate them and their variants on the ICEWS14 and Wikidata12K: (1) using only ECER to embed individual query-specific quadruples (without contextual information); (2) combining ECER and MCKR without conducting the time prediction task; (3) combining ECER and MCKR while undertaking the time prediction task (i.e., the complete ECEformer). The detailed results are listed in Table \ref{tab:abl}.

From the results of ECER, we observe that  by utilizing the Transformer architecture, our model can surpass most geoinformation-based models (such as TGeomE). This validates our viewpoint, which emphasizes sufficiently exploring the internal structure and semantic relationships within individual quadruples. Comparing the results of the ECER and ECEformer ($w/o$ TP), we observe that incorporating contextual information significantly improves the model's reasoning accuracy on Wikidata12K. However, this integration within ICEWS14 slightly diminishes performance. Specifically, ECEformer ($w/o$ TP) outperforms ECER by 2.63\% on Wikidata12K, but underperforms by 1.20\% on ICEWS14 in terms of $MRR$. This phenomenon is attributed to the finer temporal granularity of ICEWS14 compared to Wikidata12K, where the frequent historical events in the context exacerbate spurious correlations. Without the intervention of the time prediction task, the model is more prone to erroneous estimations. Comparing the results of ECEformer ($w/o$ TP) and ECEformer, we observe consistent performance enhancements on both datasets after executing the time prediction task. This is especially pronounced in ICEWS14, where the implementation of the time prediction task results in a notable augmentation of the $MRR$, surpassing the performance of the ECER. Specifically, ECEformer outperforms ECER by 1.10\%, and it outperforms ECEformer ($w/o$ TP) by 2.30\% in terms of $MRR$. These findings confirm the effectiveness of our proposed temporal information enhancement strategy in prompting the model to leverage temporal information within its context, thereby effectively mitigating the impacts of spurious correlations.

\begin{table}[t]
\caption{Sensitivity analysis of ECEformer with different number of unit layers. $N$ and $M$ represent the number of ECER units and MCKR units, respectively.}
\label{tab:layer_num}
\scalebox{1.0}{
\begin{tabular*}{\linewidth}{@{\extracolsep{\fill}}*{6}{c}}
\toprule
\multirow{2}{*}{$N$} & \multirow{2}{*}{$M$} & \multicolumn{2}{c}{ICEWS14} & \multicolumn{2}{c}{Wikidata12K} \\
&  & MRR & Hits@10 & MRR & Hits@10 \\
\midrule 
1x& \multirow{3}{*}{1} & 66.30 & 78.23 & 47.94 & 59.33\\ 
2x&& \bf{67.36} & 78.77 & 48.08 & 60.00\\
3x&& 66.95 & \bf{78.79 }& \bf{48.47} & \bf{60.24}\\
\hline
\multirow{3}{*}{1} & 1x & 66.30 & 78.23 & 47.94 & 59.33\\
& 2x & 69.01 & 78.53 & \bf{48.31} & \bf{60.19}\\
& 3x & \bf{70.30} & \bf{79.08} & 48.19 & 60.09\\
\hline\hline
1x& \multirow{3}{*}{2} & 69.04 & \bf{79.10} & \bf{48.37} & \bf{60.45}\\ 
2x& & 69.34 & 79.03 & 48.32 & 60.51\\
3x& & \bf{69.36} & 78.94 & 48.34 & 60.61\\
\hline
\multirow{3}{*}{2} & 1x & 69.04 & \bf{79.10} & \bf{48.37} & \bf{60.45}\\
& 2x & \bf{70.18} & 78.80 & 48.22 & 60.39\\
& 3x & 70.10 & 78.19 & 47.67 & 59.21\\
\bottomrule
\end{tabular*}}
\end{table}

\begin{figure}[t]
\centering
\begin{subfigure}{\linewidth}
    \centering
    \setlength{\abovecaptionskip}{-0.1cm}   
    \includegraphics[width=\linewidth]{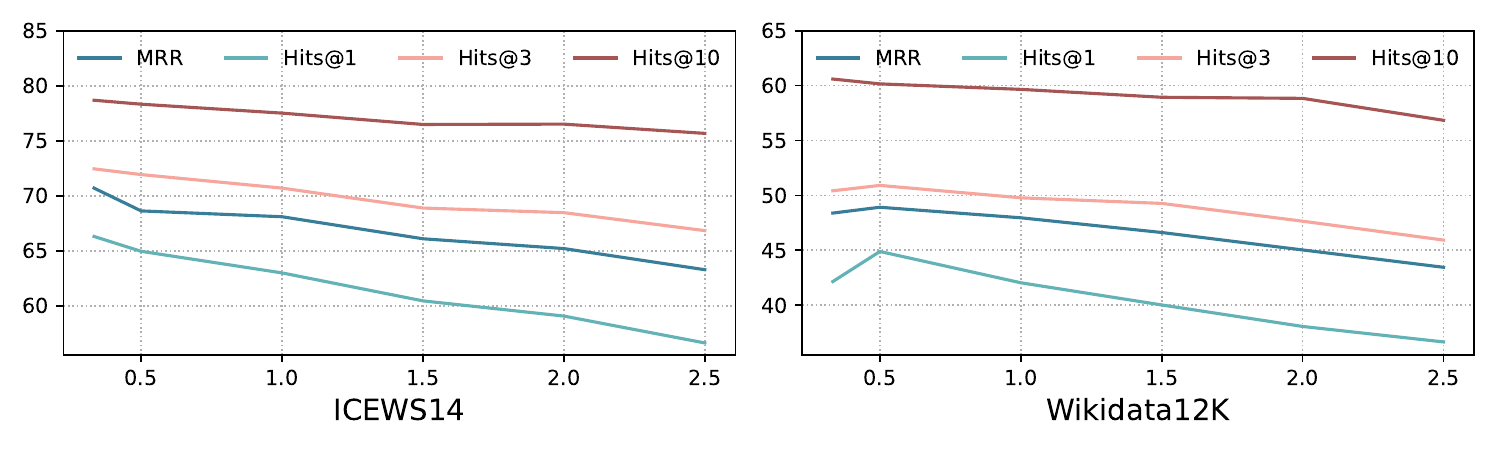} 
    \caption{Weight coefficient $\lambda$}
    \label{fig:sensi1}
\end{subfigure}
\begin{subfigure}{\linewidth}
    \centering
    \setlength{\abovecaptionskip}{-0.1cm}   
    \includegraphics[width=\linewidth]{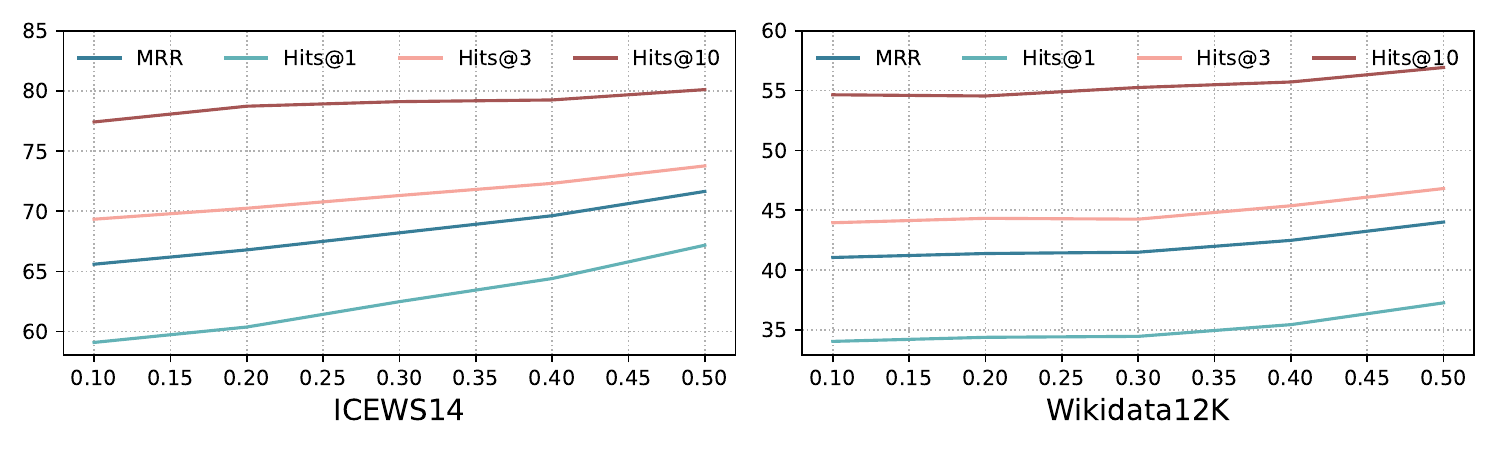} 
    \caption{Random marking rate $\gamma$}
    \label{fig:sensi2}
\end{subfigure}
\caption{Sensitivity analysis of weight coefficients $\lambda$ and random masking rate $\gamma$ on ICEWS14 and Wikidata12K.}
\label{fig:sensi}
\end{figure}

\subsection{Sensitivity Analysis}
As mentioned in \textbf{\textit{Implementation Details}}, we explore the performance fluctuations of ECEformer caused by the structural control parameters $N$ and $M$, as well as the weight coefficient $\lambda$. For structural control parameters $N$ and $M$, we set the base of one parameter to $\{1, 2\}$, while the other is adjusted in multiples, specifically to $1\times$, $2\times$, and $3\times$. As shown in Table \ref{tab:layer_num}, the results highlight the maximum value in each group in bold. Our observations reveal that different structural settings do introduce perturbations to the model, with the disturbance being relatively minor at $N:M=1:2$. Based on this finding, we configure the architecture of ECEformer as $N=3$ and $M=6$ across all datasets. For the weight coefficient $\lambda$, we record the model's performance across a range of $\{0.5, 1.0, 1.5, 2.0, 2.5\}$, and plotted these results in Figure \ref{fig:sensi1}. The performance curve trend suggests that excessively weighting the time prediction task can detrimentally affect the link prediction task, thereby reducing overall performance. Therefore, setting the weight coefficient between $0.5$ and $1.0$ offers a balanced improvement in model performance.

In addition, to investigate the impact of the masking mechanism in the time prediction task on model performance, we report the results of each batch of data under varying masking rates, with details as well as illustrated in Figure \ref{fig:sensi2}. Specifically, we employ a fixed masking rate $\gamma$ during the training phase, randomly selecting $\gamma\times\textnormal{batch\_size}$ samples per batch to mask their timestamps. We observe that as the masking rate increases, the model performance correspondingly improves. Additionally, compared to Wikidata12K, the performance of ICEWS14 exhibits greater fluctuations with varying masking rates. This observation indicates that datasets with finer temporal granularity are more sensitive to variations in masking rate. Notably, our final model is trained using a fully masked approach.

\section{Conclusion}
In this paper, we address the challenge of completing missing factual elements in TKG reasoning by introducing the ECEformer, a novel end-to-end Transformer-based reasoning model. ECEformer primarily comprises two key components: the evolutionary chain of events representation learning (ECER) and the mixed-context knowledge reasoning (MCKR). The ECER, utilizing the Transformer encoder, thoroughly explores the structural and semantic information of historical events corresponding to each branch in the evolutionary chain of events (ECE). The ECE is composed of neighborhood subgraphs of entity nodes unfolded in chronological order. The MCKR based on MLP relies on the channel and patch mixing strategy to facilitate the learning of contextual information interactions within the ECE. Moreover, an additional time prediction task and a time masking mechanism are employed to force the model to assimilate temporal information from the context. Comparative experiments with state-of-the-art methods validate the superiority of our approach. Furthermore, additional ablation studies demonstrate the effectiveness of each module proposed in our ECEformer.

\begin{acks}
  This research was supported by National Science and Technology Major Project (2022ZD0119204), National Science Fund for Distinguished Young Scholars (62125601), National Natural Science Foundation of China (62076024, 62172035).
\end{acks}

\bibliographystyle{ACM-Reference-Format}
\bibliography{sample-base}


\end{document}